\documentclass{article}
\usepackage{ijcai_style/ijcai20}

\usepackage{times}

\usepackage{soul}
\usepackage{url}
\usepackage[hidelinks]{hyperref}
\usepackage[utf8]{inputenc}
\usepackage[small]{caption}
\usepackage{graphicx}
\usepackage{amsmath}
\usepackage{booktabs}
\usepackage{nicefrac}       
\usepackage{graphics}
\usepackage{algorithm}
\usepackage{algorithmic}
\usepackage{subcaption}
\usepackage{todonotes}

\usepackage{latexsym} 

\title{Reward Prediction Error as an Exploration Objective in Deep RL}

\author{
Riley Simmons-Edler$^{1,2}$\footnote{Contact Author}\and
Ben Eisner$^2$\and
Daniel Yang$^2$\and
Anthony Bisulco$^2$\and
Eric Mitchell$^{2,3}$\and
Sebastian Seung$^2$ \And
Daniel Lee$^2$\\
\affiliations
$^1$Princeton University\\
$^2$Samsung AI Center NYC\\
$^3$Stanford University\\
\emails
rileys@cs.princeton.edu
}

\begin{document}

\maketitle

\begin{abstract}
A major challenge in reinforcement learning is \emph{exploration}, when local dithering methods such as $\epsilon$-greedy sampling are insufficient to solve a given task. Many recent methods have proposed to intrinsically motivate an agent to seek novel states, driving the agent to discover improved reward. However, while state-novelty exploration methods are suitable for tasks where novel observations correlate well with improved reward, they may not explore more efficiently than $\epsilon$-greedy approaches in environments where the two are not well-correlated. In this paper, we distinguish between exploration tasks in which seeking novel states aids in finding new reward, and those where it does not, such as goal-conditioned tasks and escaping local reward maxima. We propose a new exploration objective, maximizing the reward prediction error (RPE) of a value function trained to predict extrinsic reward. We then propose a deep reinforcement learning method, QXplore, which exploits the temporal difference error of a Q-function to solve hard exploration tasks in high-dimensional MDPs. We demonstrate the exploration behavior of QXplore on several OpenAI Gym MuJoCo tasks and Atari games and observe that QXplore is comparable to or better than a baseline state-novelty method in all cases, outperforming the baseline on tasks where state novelty is not well-correlated with improved reward.

\end{abstract}

\section{Introduction}
\label{sec:intro}


In recent years deep reinforcement learning (RL) algorithms have demonstrated impressive performance on tasks such as playing video games and controlling robots \cite{Mnih2015,kalashnikov2018qtopt}. However, successful training for such cases typically requires both a well-shaped reward function, where the RL agent can sample improved trajectories through simple dithering exploration such as $\epsilon$-greedy sampling, and the ability to collect many (hundreds of thousands to millions) of trials. Satisfying these preconditions often requires large amounts of domain-specific engineering. In particular, reward function design can be unintuitive, may require many iterations of design, and in some domains such as robotics can be physically impractical to implement.


The field of \emph{exploration} methods in RL seeks to address the difficulties of reward design by allowing RL agents to learn from \emph{unshaped} reward functions. Unshaped functions (for example, a reward of 1 when an object is moved to a target, and 0 otherwise) are usually much easier to design and implement than dense well-shaped reward functions. However, it is hard for standard RL algorithms to discover good policies on unshaped reward functions, and they may learn very slowly, if at all.


While substantial work has been conducted on designing general exploration strategies for high-dimensional Markov Decision Processes (MDPs) with sparse reward functions, few studies have distinguished between different types of tasks requiring exploration, particularly in terms of which signals in each MDP are useful for discovering new sources of reward. In this work, we consider three types of exploration challenges in particular: solving mazes, learning goal conditioning relationships, and escaping local reward maxima.

Many classical exploration tasks can be described well as \textbf{mazes}. For example, discovering the single rewarding state in a sparse reward environment, or navigating a precise series of obstacles in order to play a game. Qualitatively, the agent must search for the exit to the maze (reward), receives little or no reward before finding it, and has no learned priors. In the limit any RL task can be seen as a maze (such as by treating a single optimal trajectory as the ``exit''), but such a treatment is often intractable for large MDPs.

Related but distinct are \textbf{goal conditioned tasks}. Here, the reward function is conditioned on a non-static goal specified by the environment and discovered through interaction. For example, a robot that must move an object to a set of coordinates, which differ for each episode. The agent must learn how the observation and reward are conditioned on the goal, which is made significantly harder when the underlying reward function is sparse and unshaped. Unlike a maze, correlations between observation and goal/reward provide additional information an agent can use to solve the problem.

Lastly, in a poorly-shaped reward function there may exist local maxima in the space of trajectories, where an agent cannot discover an improved policy through local exploration and must deliberately sample suboptimal trajectories to \textbf{escape local maxima}. Here, the contours of the reward function can provide information on what directions of exploration might be informative, even if exploring them does not immediately maximize reward.



This distinction is important because both goal conditioning and local maxima introduce additional information about the task that mazes do not contain \textemdash{} In a maze-like environment, discovering new states is explicitly linked with discovering new reward signals. Goal conditioning can provide hints as to what states are and are not rewarding (and when) through correlations in the observation. Similarly, local reward maxima are embedded within a dense reward function which provides correlations between each observation and the reward that results, and discovering this relationship may lead to improved reward. Each of these problems can be intractable using naive exploration (depending on the severity of the problem), but each in turn provides some signal that can be used to solve it.
For example, goal-conditioning relationships can also be learned by goal-driven RL methods such as Hindsight Experience Replay \cite{andrychowicz2017her}, which by assuming the presence of a goal can learn much faster and more sample-efficiently on that class of problems.




In this paper, we propose the use of reward prediction error, specifically the Temporal-Difference Error (TD-Error) of a value function, to direct exploration in MDPs that contain Goal Conditioning and Local Maxima Escape problems but do not have a strong correlation between reward discovery and state novelty. To facilitate the use of this objective in a deep reinforcement learning setting for high-dimensional MDPs, we introduce QXplore, a new deep RL exploration formulation that seeks novelty in the predicted reward landscape instead of novelty in the state space. QXplore exploits the inherent reward-space signal from TD-error in value-based RL, and directly promotes visiting states where the current understanding of reward dynamics is poor. In the following sections, we describe QXplore for continuous MDPs and demonstrate its utility for efficient learning on a variety of complex benchmark environments showcasing different exploration cases.

\section{Related Work}
\label{sec:related_work}

Of the exploration methods proposed for deep RL settings, the majority provide some state-novelty objective that incentivizes an agent to explore novel states or transition dynamics. A simple approach consists of explicitly counting how many times each state has been visited, and acting to visit rarely explored states. This approach can be useful for small MDPs, but often performs poorly in high-dimensional or continuous state spaces. However, several recent works \cite{tang2017exploration,bellemare2016unifying,fu2017ex2} using count-like statistics have shown success on benchmark tasks with complex state spaces.

Another approach to environment novelty learns a model of the environment's transition dynamics and considers novelty as the error of the model in predicting future states or transitions. This exploration method relies on the assumption that any new state that can be predicted in advance is equivalent to some previously seen state in its effect on reward. Predictions of the transition dynamics can be directly computed \cite{Pathak2017curiosity,Stadie2015IncentivizingEI}, or related to an information gain objective on the state space, as described in VIME \cite{houthooft2016vime} and EMI \cite{kim2019emi}. 

Several exploration methods have recently been proposed that capitalize on the function approximation properties of neural network to recognize novel states. Random network distillation (RND) trains a function to predict the output of a randomly-initialized neural network from an input state, and uses the approximation error as a reward bonus for a separately-trained RL agent \cite{burda2018rnd}. Similarly, DORA \cite{fox2018dora} trains a network to predict zero on observed states and deviations from zero are used to indicate unexplored states.

These methods have been shown to perform well on maze-solving exploration tasks such as the Atari game \texttt{Montezuma's Revenge}, where maximizing reward (game score) requires visiting each room of the game, which also maximizes the diversity of states and observations experienced. However, evaluating these methods on tasks where novelty does not correlate highly with reward, such as on other Atari games, shows little improvement over $\epsilon$-greedy \cite{taiga2020on}. 

Reward prediction error has been previously used for exploration in a few cases. Previous works described using reward misprediction and model prediction error for exploration \cite{schmidhuber1991adaptive,thrun1992active}. However, these works were primarily concerned with model-building and system-identification in small MDPs, and used single-step reward prediction error rather than TD-error. Later, TD-error was used as a negative signal to constrain exploration to focus on states that are well understood by the value function for safe exploration \cite{Gehring2013smartexploration}. Related to maximizing TD-error is maximizing the variance or KL-divergence of a posterior distribution over MDPs or Q-functions, which can be used as a measure of uncertainty about rewards \cite{fox2018dora,osband2018randomized}. Posterior uncertainty over Q-functions can be used for information gain in the reward or Q-function space, but posterior uncertainty methods have thus-far largely been used for local exploration as an alternative to dithering methods such as $\epsilon$-greedy sampling, though \cite{osband2018randomized} do apply posterior uncertainty to \texttt{Montezuma's Revenge} and other exploration tasks in the Atari game benchmark.

\section{Preliminaries}
\label{sec:preliminaries}

We consider RL in the terminology of \cite{sutton1998rl}, in which an agent seeks to maximize reward in a Markov Decision Process (MDP).
An MDP consists of states $s \in \mathcal{S}$, actions $a \in \mathcal{A}$, a state transition function $S:\mathcal{S} \times \mathcal{A} \times \mathcal{S} \rightarrow [0,1]$ giving the probability of moving to state $s_{t+1}$ after taking action $a_t$ from state $s_t$ for discrete timesteps $t \in {0,...,T}$.
Rewards are sampled from reward function $r:\mathcal{S}\times \mathcal{A} \rightarrow R$.
An RL agent has a policy $\pi(s_t,a_t) = p(a_t | s_t)$ that gives the probability of taking action $a_t$ when in state $s_t$.
The agent aims to learn a policy to maximize the expectation of the time-decayed sum of reward $R_{\pi}(s_0) = \sum_{t=0}^T\gamma^tr(s_t,a_t)$ where $a_t\sim \pi(s_t,a_t)$.

A value function $V_{\theta}(s_t)$ with parameters $\theta$ is a function which computes $V_{\theta}(s_t) \approx R_{\pi}(s_t)$ for some policy $\pi$. Temporal difference (TD) error $\delta_t$ measures the bootstrapped error between the value function at the current timestep and the next timestep as 
\begin{equation}
    \label{eq:td_definition}
    \delta_t = V_{\theta}(s_t) - (r(s_t, a_t \sim \pi(s_t)) + {\gamma}V_{\theta}(s_{t+1})).
\end{equation}
A Q-function is a value function of the form $Q(s_t,a_t)$, which computes $Q(s_t, a_t)~=~r(s_t, a_t) + {\gamma}\cdot\text{max}_{a'}Q(s_{t+1}, a')$, the expected future reward assuming the optimal action is taken at each future timestep. An approximation to this optimal Q-function $Q_\theta$ with some parameters $\theta$ may be trained using a mean squared TD-error objective $L_{Q_{\theta}} = ||Q_{\theta}(s_t,a_t) - (r(s_t,a_t) + {\gamma}\cdot \text{max}_{a'}Q'_{\theta'}(s_{t+1}, a'))||^2$ given some target Q-function $Q'_{\theta'}$, commonly a time-delayed version of $Q_{\theta}$ \cite{Mnih2015}. Extracting a policy $\pi$ given $Q_{\theta}$ amounts to computing $\text{argmax}_aQ_{\theta}(s_t,a)$. 











\section{QXplore: TD-Error as Reward Signal}
\label{sec:method}


\subsection{TD-error Objective}
\label{subsec:tderr}

We first discuss why and how TD-error can be used as an exploration signal in deep RL settings on the classes of MDPs discussed above. Many Deep RL methods maintain a value function, typically a Q function, which in off-policy settings is bootstrapped to approximate the true Q function of the optimal policy. During the course of training, this Q function will naturally contain inaccuracies such that there is nontrivial Bellman error for certain  $s, a, s', r$ tuples. Intuitively, these errors indicate that the current estimate of the Q function does not correctly model the reward dynamics of the MDP per Bellman optimality. Therefore, an exploration method that prioritizes seeking out regions of the environment where the Q-function is inaccurate could aid an off-policy method in discovering novel sources of reward and propagating those improvements through the Q function.


Given a Q function with parameters $\theta$ and $\delta_t$ we define our exploration signal for a given state-action-next-state tuple as:
\begin{align}
\label{eq:rx}
r_{x, \theta}(s_t,a_t,s_{t+1}) = |\delta_t| = |Q_{\theta}(s_t,a_t) - \nonumber\\ 
(r_{\text{E}}(s_t, a_t) + {\gamma}\text{max}_{a'}Q'_{\theta'}(s_{t+1},a'))|
\end{align}

for some extrinsic reward function $r_{\text{E}}$ and target Q-function $Q'_{\theta}$. Notably, we use the absolute value of the TD rather than signed TD, as this is necessary to harness network extrapolation error in sparse reward environments.


Intuitively, a policy maximizing the expected sum of $r_x$ for a fixed Q function will sample trajectories where $Q_{\theta}$ does not have an accurate estimate of the future rewards it will experience.
This is useful for exploration because $r_x$ will be large not only for state-action pairs producing unexpected reward, but for all state-action pairs leading to such states, providing a denser exploration reward function and allowing for longer-range exploration.

\begin{algorithm}[tb]
   \caption{QXplore for Continuous Actions}
   \label{alg:qxplore}
\begin{algorithmic}
   \STATE {\bfseries Input:} MDP $S$, Q-function $Q_{\theta}$ with target $Q'_{\theta'}$, $Q_x$ function $Q_{x,\phi}$ with target $Q'_{x, \phi'}$, replay buffers $\mathcal{Z}_Q$ and $\mathcal{Z}_{Q_x}$, batch size $B$ and sampling ratios $\mathcal{R}_Q$ and $\mathcal{R}_{Q_x}$, CEM policies $\pi_Q$ and $\pi_{Q_x}$, time decay parameter $\gamma$, soft target update rate $\tau$, and environments $E_Q$, $E_{Q_x}$
   \WHILE{not converged}
   \STATE Reset $E_Q$, $E_{Q_x}$
   \WHILE{$E_Q$ and $E_{Q_x}$ are not done}
   \vskip 1mm
   \STATE \textbf{Sample environments}
   \STATE $\mathcal{Z}_Q \leftarrow (s,a,r,s') \sim \pi_Q | E_Q$
   \STATE $\mathcal{Z}_{Q_x} \leftarrow (s,a,r,{s'}) \sim \pi_{Q_x} | E_{Q_x}$
   \vskip 1mm
   \STATE \textbf{Sample minibatches for $Q_{\theta}$ and $Q_{x, \phi}$}
   \STATE $ (s_Q,a_Q,r_Q,{s'}_Q) \leftarrow B*\mathcal{R}_Q$ samples from $\mathcal{Z}_Q$ and $B*(1 - \mathcal{R}_{Q})$ samples from $\mathcal{Z}_{Q_x}$
   \STATE $ (s_{Q_x},a_{Q_x},r_{Q_x},{s'}_{Q_x}) \leftarrow B*\mathcal{R}_{Q_x}$ samples from $\mathcal{Z}_{Q_x}$ and $B*(1 - \mathcal{R}_{Q_x})$ samples from $\mathcal{Z}_Q$
   \vskip 1mm
   \STATE \textbf{Train}
   \STATE $r_{x,\theta} \leftarrow |Q_{\theta}(s_{Q_x}, a_{Q_x}) - $
   \STATE $(r_{Q_x} + {\gamma}Q'_{\theta'}({s'}_{Q_x}, \pi_Q({s'}_{Q_x})))|$
   \STATE $L_Q \leftarrow ||Q_{\theta}(s_Q, a_Q) - (r_Q + {\gamma}Q'_{\theta'}({s'}_Q, \pi_Q({s'}_Q)))||^2$
   \vskip 0mm 
   \STATE $L_{Q_x} \leftarrow ||Q_{x,\phi}(s_{Q_x}, a_{Q_x}) - $
   \STATE $(r_{x,\theta} + {\gamma}Q'_{x,\phi'}({s'}_{Q_x}, \pi_{Q_x}({s'}_{Q_x})))||^2$
   \STATE Update $\theta \propto L_Q$
   \STATE Update $\phi \propto L_{Q_x}$
   \STATE $\theta' \leftarrow (1 - \tau)\theta' + \tau\theta$
   \STATE $\phi' \leftarrow (1 - \tau)\phi' + \tau\phi$
   \ENDWHILE
   \ENDWHILE
   
\end{algorithmic}
\end{algorithm}

\subsection{$Q_x$: Learning a Q-Function to Maximize TD-error}
\label{subsec:qx}

Now that we have defined a TD-error exploration formulation, we must ask, how should we maximize it? If we treat this signal as a reward function, $r_x$ can be used to generate a new MDP where the reward function is replaced by $r_x$, and thus generally can be solved via any RL algorithm. For practical purposes, we choose to train a second Q-function to maximize $r_x$, as allows the entire algorithm to be trained off-policy and for the two Q-functions to share replay data. Additionally, this allows us to use the original Q function $Q_{\theta}$ as an exploitation policy at inference time, avoiding the need to trade off between exploration and exploitation because the Q function estimates are not directly affected by $r_x$ values.

In our formulation, which we call QXplore, we define a Q-function, $Q_{x,\phi}(s,a)$ with parameters $\phi$, whose reward objective is $r_x$. We train $Q_{x,\phi}$ using the standard bootstrapped loss function
\begin{align}
    \label{eq:qx}
L_{Q_{x, \phi}} = ||Q_{x,\phi}(s_t,a_t) - (r_x(s_t,a_t,s_{t+1}) + \nonumber\\
{\gamma}\text{max}_{a'}Q'_{x,\phi'}(s_{t+1}, a'))||^2 .
\end{align}

\begin{figure}[t]
\begin{center}
\centerline{\includegraphics[width=\columnwidth]{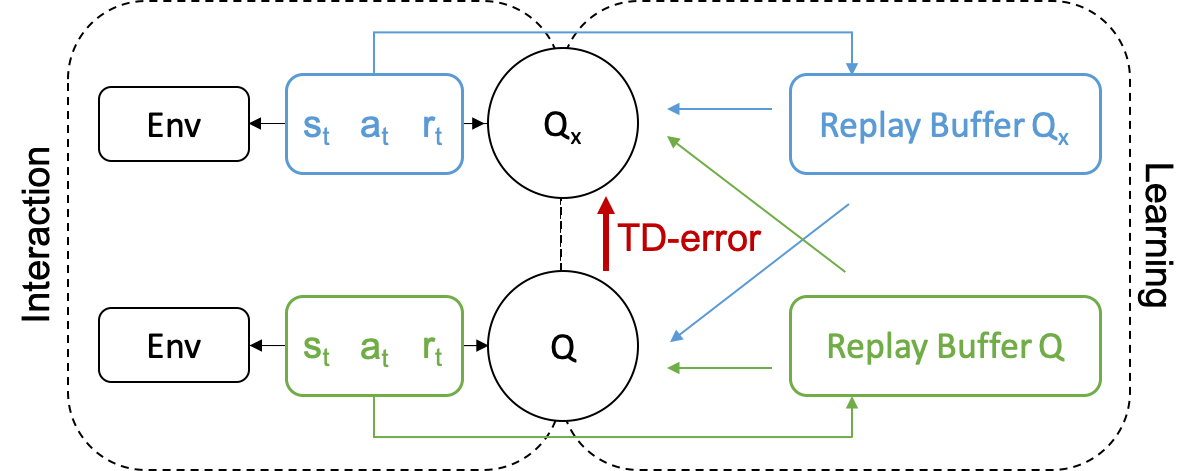}}
\caption{Method diagram for QXplore. We define two Q-functions which sample trajectories from their environment and store experiences in separate replay buffers. $Q$ is a standard state-action value-function, whereas $Q_x$'s reward function is the unsigned temporal difference error of the current Q on data sampled from both replay buffers. A policy defined by $Q_x$ samples experiences that maximize the TD-error of $Q$, while a policy defined by $Q$ samples experiences that maximize discounted reward from the environment.}
\label{fig:method}
\end{center}
\end{figure}

The two Q-functions, $Q_{\theta}$ and $Q_{x}$, are trained in parallel, sharing replay data so that $Q_{\theta}$ can learn to exploit sources of reward discovered by $Q_{x}$ and so that $Q_{x}$ can better predict the TD-errors of $Q_{\theta}$. Since the two share data, $\pi_{Q_x}$ acts as an adversarial teacher for $Q_{\theta}$, sampling trajectories that produce high TD-error under $Q_{\theta}$ and thus provide novel information about the reward landscape. A similar adversarial sampling scheme was used to train an inverse dynamics model by \cite{hong2019adversarial}, and \cite{colas2018gep} use separate goal-driven exploration and reward maximization phases for efficient learning. However, to our knowledge adversarial sampling policies have not previously been used for exploration. To avoid off-policy stability issues due to the different reward objectives, we sample a fixed ratio of experiences collected by each policy for each training batch.
Our full method is described for the continuous-action domain in Algorithm \ref{alg:qxplore} and a schematic of the method is shown in Figure \ref{fig:method}.

\subsection{State Novelty from Neural Network Function Approximation Error}
\label{subsec:0_predict}
\begin{figure}[t]
\begin{center}
\includegraphics[width=0.99\columnwidth]{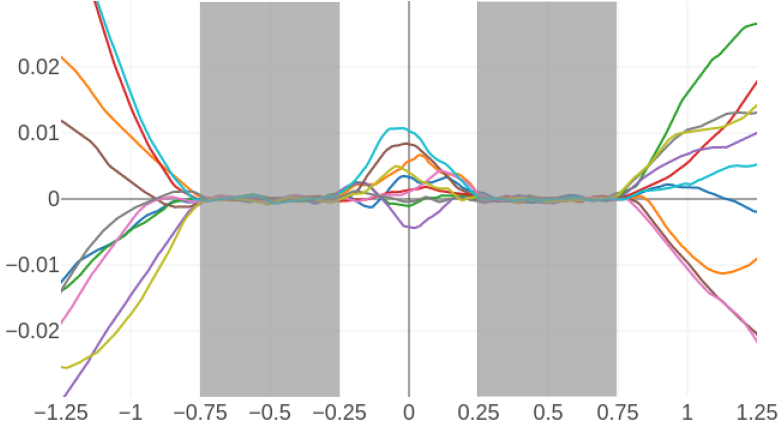}
\caption{A neural network trained to predict a constant value does not interpolate or extrapolate well outside its training range, which can be exploited for exploration. Predictions of 3-layer MLPs of 256 hidden units per layer trained to imitate $f(x) = 0$ on $R \rightarrow R$ with training data sampled uniformly from the range $[-0.75,-0.25]\cup[0.25,0.75]$. Each line is the final response curve of an independently trained network once its training error has converged (MSE $<$ 1e-7).}
\label{fig:0pred}
\end{center}
\end{figure}

A key question in using TD-error for exploration is what happens when the reward landscape is flat? Theoretically, in the case that $\forall(s,a),r(s,a) = C$ for some constant $C \in R$, an optimal Q-function which generalizes perfectly to unseen states will, in the infinite time horizon case, simply output $\forall(s,a), Q^{\star}(s,a) = \sum_{t=0}^{\infty}C\gamma^t$. This results in a TD-error of 0 everywhere and thus no exploration signal. However, using neural network function approximation, we find that perfect generalization to unseen states-action pairs does not occur, and in fact observe in Figure \ref{fig:0pred} that the distance of a new datum from the training data manifold correlates with the magnitude of the network output's deviation from $\sum_{t=1}^{\infty}C\gamma^t$ and thus with TD-error. As a result, in the case where the reward landscape is flat TD-error exploration converges to a form of state novelty exploration. This property of neural network function approximation has been used by several previous exploration methods to good effect, including RND \cite{burda2018rnd} and DORA \cite{fox2018dora}. In particular, the exploration signal used by RND (extrapolation error from fitting the output of a random network) should be analogous to $r_x$ (extrapolation error from fitting a constant value), meaning we should expect to perform comparably to RND when no extrinsic reward exists. 


\section{Experiments}
\label{sec:experiments}
We describe here the results of experiments to demonstrate the effectiveness of QXplore on continuous control and Atari benchmark tasks. We also compare to results on \texttt{SparseHalfCheetah} from several previous publications. Finally, we discuss several ablations to QXplore to demonstrate that all components of the method improve performance. 

We compare QXplore primarily with a related state of the art state novelty-based method, RND \cite{burda2018rnd}, and with $\epsilon$-greedy sampling as a simple baseline.  Each method is implemented in a shared code base on top of TD3/dueling double deep Q-networks for the continuous/discrete action case \cite{fujimoto2018td3,wang2016dueling}. For experiments in continuous control environments, we implement and use a nonparametric cross-entropy method policy, previously described as more robust to hyperparameter variance, with the same architecture and hyperparameters as prior work \cite{simmons2019q,kalashnikov2018qtopt}. We experimented with a variant using DDPG-style parametric policies \cite{Lillicrap2015} for both $Q_{\theta}$ and $Q_x$ , but found preventing $Q_{\theta}$'s policy from converging to poor local maxima difficult, consistent with previously reported stability issues in that class of algorithms \cite{simmons2019q,Islam2017}. For all experiments, we set the data sampling ratios of $Q_{\theta}$ and $Q_x$, $\mathcal{R}_Q$ and $\mathcal{R}_{Q_x}$ respectively, at 0.75, the best ratio among a sweep of ratios 0.0, 0.25, 0.5, and 0.75 on \texttt{SparseHalfCheetah}. For continuous control tasks, we used a learning rate of 0.0001 for both Q-functions, the best among all paired combinations of 0.01, 0.001, and 0.0001, and fully-connected networks of two hidden layers of 256 neurons to represent each Q-function, with no shared parameters. For Atari benchmark tasks, we used the dueling double deep Q-networks architecture and hyperparameters described by \citeauthor{wang2016dueling}.


\subsection{Experimental Setup}
\label{subsec:setup}

We benchmark on five continuous control tasks using the MuJoCo physics simulator that each require exploration due to sparse or unshaped rewards. First, the \texttt{SparseHalfCheetah} task originally proposed by VIME \cite{houthooft2016vime}. This task requires an agent to move 5 units (several hundred timesteps of actions) forward to receive reward, receiving 0 reward otherwise, and is maze-like in this regard. Next, we benchmark on three goal-directed OpenAI gym tasks, \texttt{FetchPush}, \texttt{FetchSlide} and \texttt{FetchPickAndPlace}, originally proposed in HER \cite{andrychowicz2017her}. Lastly, we test a variant of \texttt{SparseHalfCheetah} that we refer to as \texttt{LocalMaxEscape} where a local reward maximum has been introduced \textemdash{} the agent receives 0 reward for every timestep it is between -1 and 1 units from the origin, -1 reward if it moves outside that range, but 100 reward per timestep if it moves 5 units forward, similar to \texttt{SparseHalfCheetah}. We chose these tasks as they are challenging exploration problems highlighting the different cases we are interested in that are relatively simple to control, but still involve large continuous state spaces and continuous actions. Guided by a recent study suggesting that exploration in the Atari game benchmark suite doesn't improve performance on most tasks \cite{taiga2020on}, we evaluated on a pair of ``hard'' exploration games, \texttt{Venture} and \texttt{Gravitar}, as well as an easy game, \texttt{Pong}, to show that QXplore can also function in this very different domain. We ran five random seeds for each experiment and plot the mean and plus/minus 1 standard deviation bounds for each set of runs, applying a Gaussian filter to each mean/stdev for readability.

\subsection{Exploration Benchmark Performance}

\begin{figure*}[t]

    \begin{center}
    \begin{subfigure}[b]{0.24\textwidth}
        \centering
        \includegraphics[width=\textwidth]{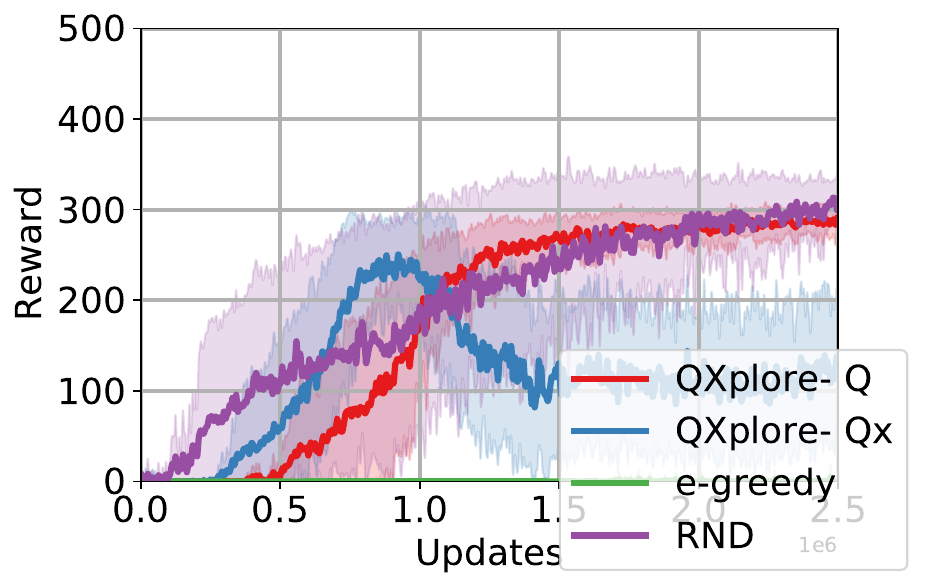}
        \caption{\label{fig:figa} SparseHalfCheetah-v1}
    \end{subfigure}
    \begin{subfigure}[b]{0.24\textwidth}
        \includegraphics[width=\textwidth]{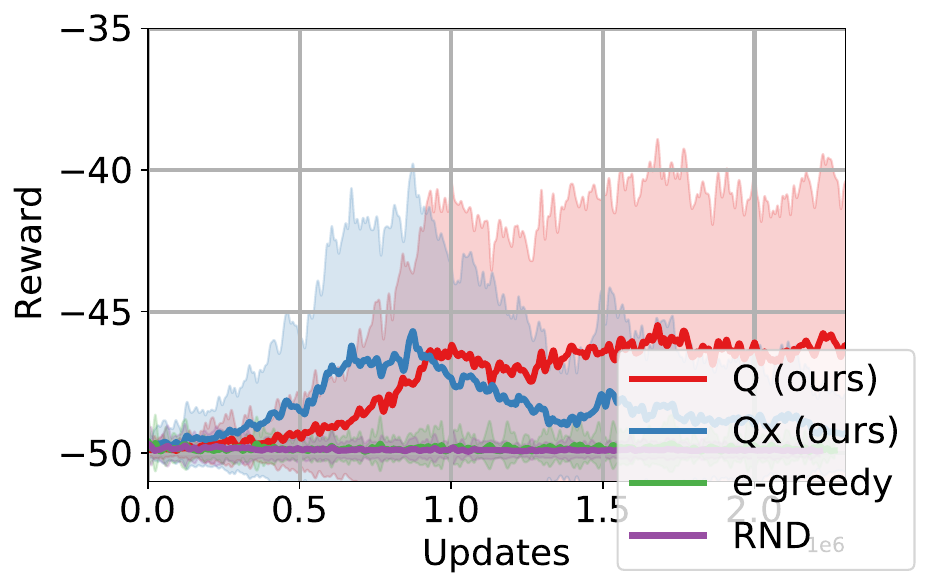}
        \caption{\label{fig:figb} FetchSlide-v1}
    \end{subfigure}
   \begin{subfigure}[b]{0.24\textwidth}
        \centering
        \includegraphics[width=\textwidth]{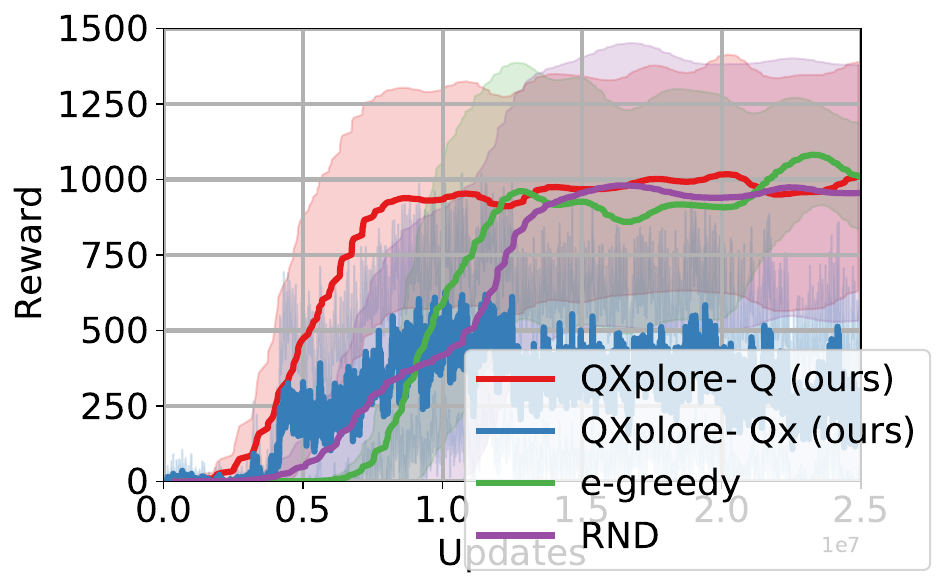}
        \caption{\label{fig:figc} Venture-v0}
    \end{subfigure}
   \begin{subfigure}[b]{0.24\textwidth}
        \centering
        \includegraphics[width=\textwidth]{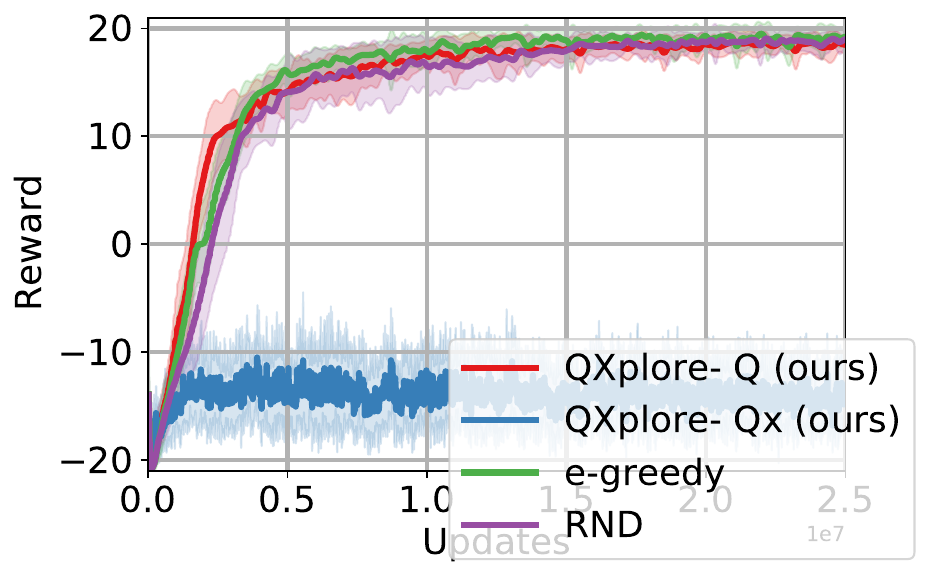}
        \caption{\label{fig:figd} Pong-v0}
    \end{subfigure}

   \medskip

   \begin{subfigure}[b]{0.24\textwidth}
        \centering
        \includegraphics[width=\textwidth]{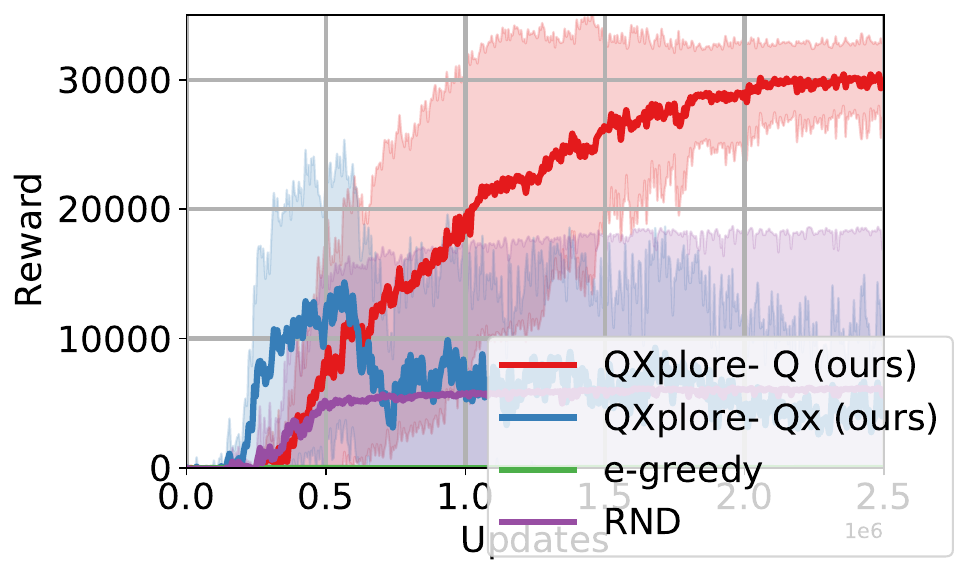}
        \caption{\label{fig:fige} LocalMaxEscape-v1}
    \end{subfigure}   
   \begin{subfigure}[b]{0.24\textwidth}
        \centering
        \includegraphics[width=\textwidth]{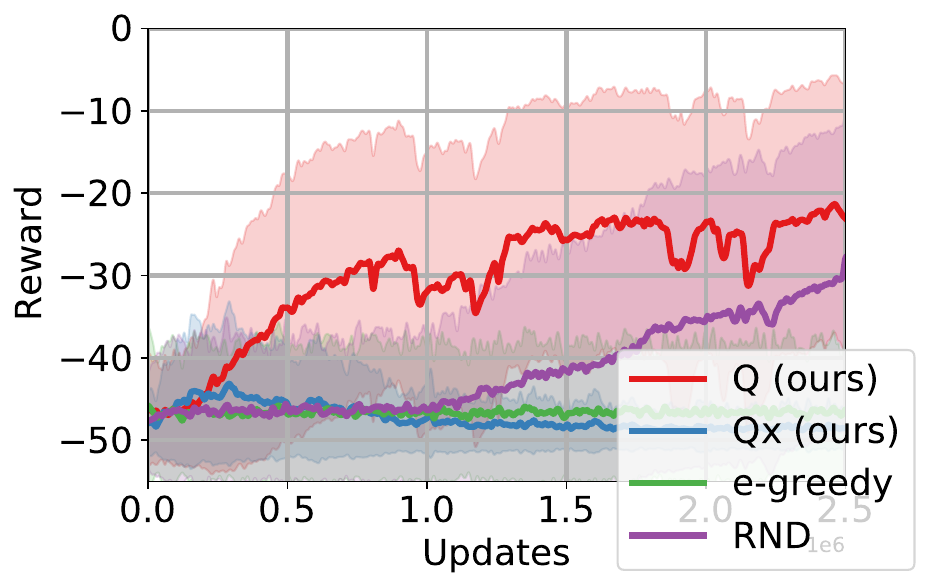}
        \caption{\label{fig:figf} FetchPush-v1}
    \end{subfigure}
    \begin{subfigure}[b]{0.24\textwidth}
        \centering
        \includegraphics[width=\textwidth]{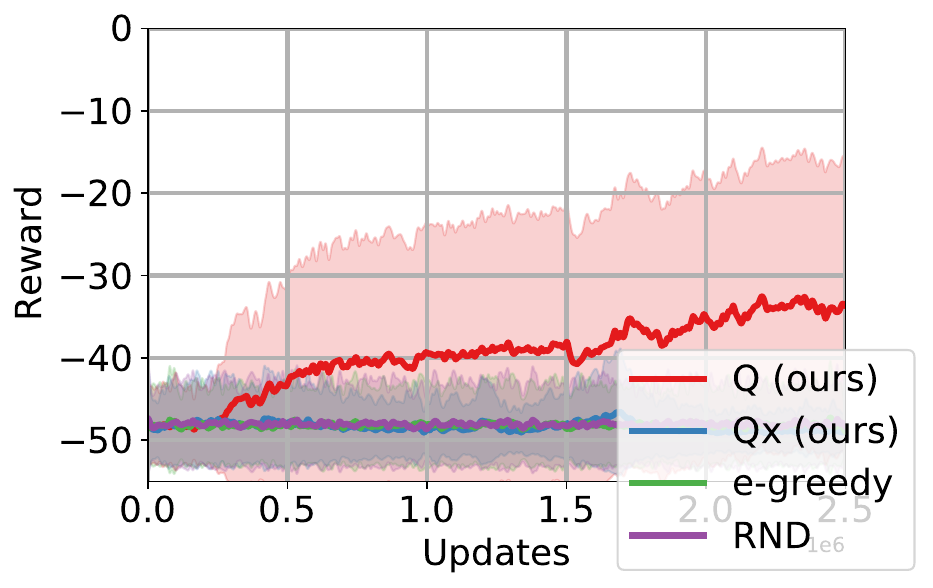}
        \caption{\label{fig:figg} FetchPickAndPlace-v1}
    \end{subfigure}
    \begin{subfigure}[b]{0.24\textwidth}
        \centering
        \includegraphics[width=\textwidth]{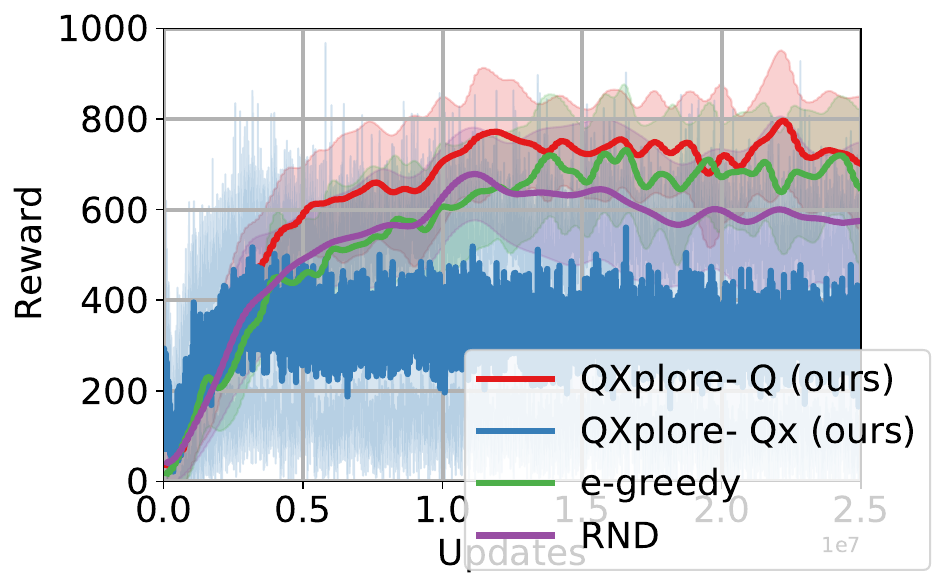}
        \caption{\label{fig:figh} Gravitar-v0}
    \end{subfigure}


\caption{Performance of QXplore compared with RND and $\epsilon$-greedy sampling. QXplore outperforms RND and $\epsilon$-greedy on the Fetch tasks and in escaping local maxima, while performing comparably on maze solving tasks and non-exploration tasks. ``QXplore-Q'' indicates the performance of our exploitation Q-function, while ``QXplore-Qx'' indicates the performance of our exploration Q-function, whose objective does not directly maximize reward but which may lead to high reward regardless.}

\label{fig:benchmarks}
\end{center}
\end{figure*}

\begin{table}
    \centering
    \begin{tabular}{|c|c|c|c|c|}
        \hline
        \textbf{Episodes until} & & & &  \\
        \textbf{mean reward of}& \textbf{QXplore} & \textbf{VIME} & \textbf{EX2} & \textbf{EMI}\\\hline
        \textbf{50} & 3000 & 10000* & 4740* & \textbf{2580}* \\\hline
        \textbf{100} & \textbf{3400} & x* & 6180* & 4520* \\\hline
        \textbf{200} & \textbf{4000} & x* & x* & 8440* \\\hline
        \textbf{300} & \textbf{10000} & x* & x* & x* \\\hline
    \end{tabular}
    \caption{Number of episodes required to reach mean reward milestones on \texttt{SparseHalfCheetah} for several methods. QXplore reaches higher rewards than previously published results. Results marked with ``*'' are previously published numbers.
    Results marked with ``x'' indicate that the mean reward was not achieved.}
    \label{table:comparison}
\end{table}
\label{subsec:benchmarks}

We show the performance of each method on each task in Figure \ref{fig:benchmarks}. QXplore performs comparable to RND on the \texttt{SparseHalfCheetah} task, in line with our expectation, but performs much better comparatively on the \texttt{Fetch} tasks \textemdash{} only on \texttt{FetchPush}, the easiest task, did RND find non-random reward. We believe this is because TD-error drives exploration behavior that helps the agent to uncover the goal-conditioning relationship, whereas state novelty is goal-agnostic and does not aid in discovery of the relationship. QXplore also strongly outperformed RND on \texttt{LocalMaxEscape}, as negative rewards far from the origin increase TD error and drive rapid discovery of the global optimum.

To validate our performance and sample efficiency, we compare QXplore to previously published \texttt{SparseHalfCheetah} performance numbers in Table \ref{table:comparison}. Because to our knowledge no previous work has evaluated off-policy Q-learning based methods on \texttt{SparseHalfCheetah}, we compare to previous methods built on top of TRPO \cite{schulman2015trust}. Due to the difference in baseline algorithms, we compare the number of episodes of interaction required to reach a given level of reward, though QXplore was not intended to be performant with respect to this metric.
While some decrease in sample efficiency is expected due to differing baseline methods (TRPO \cite{schulman2015trust} versus TD3 \cite{fujimoto2018td3}), compared to the results reported by \citeauthor{kim2019emi} for EMI \cite{kim2019emi} and EX2 \cite{fu2017ex2}, and by \citeauthor{houthooft2016vime} for VIME \cite{houthooft2016vime} on the \texttt{SparseHalfCheetah} task, QXplore reaches almost every reward milestone faster, and achieves a peak reward (300) not achieved by any previous method. This shows that off-policy Q-learning combined with TD-error exploration can result in sample efficient as well as flexible exploration.

We also implemented a continuous-control adaptation of DORA \cite{fox2018dora} and tested it on \texttt{SparseHalfCheetah}. DORA performed poorly, possibly because it was not intended for use with continuous action spaces, and thus we did not test it on other tasks.

As a comparison to a published off-policy Q-learning exploration method, we compared to GEP-PG \cite{colas2018gep}, which used separate exploration and exploitation phases similar to QXplore. We downloaded the author's implementation (built on top of DDPG) and tested it on \texttt{SparseHalfCheetah} using the parameters for the \texttt{HalfCheetah-v2} task it was originally tested on. GEP-PG reached a validation reward of 120.2 after 4000 episodes, broadly comparable to our QXplore and RND implementations.

Finally, while the main focus of our evaluation is on continuous control tasks, we also evaluated QXplore on several games in the Atari Arcade Learning Environment \cite{bellemare2013arcade} to verify that QXplore extends to tasks with image observations and discrete action spaces. We implemented QXplore and RND on top of dueling double DQN \cite{wang2016dueling}, using hyperparameters and network architectures from the Dopamine implementation of DQN \cite{castro18dopamine}. We show the results in Figure \ref{fig:benchmarks} after 25 million training steps. Based on the findings of \cite{taiga2020on}, we did not expect to improve significantly compared to the baseline in this domain. Indeed, we find that QXplore performs comparably to the baseline and RND implementations on \texttt{Pong} and \texttt{Gravitar}, while outperforming them modestly on \texttt{Venture}, where QXplore converges faster, perhaps due to $Q_x$ focusing on reward-adjacent states more than $\epsilon$-greedy or RND. 

\subsection{Ablations}
\label{subsec:ablations}
\begin{figure}[t]
\begin{center}

\centering
\includegraphics[width=0.49\columnwidth]{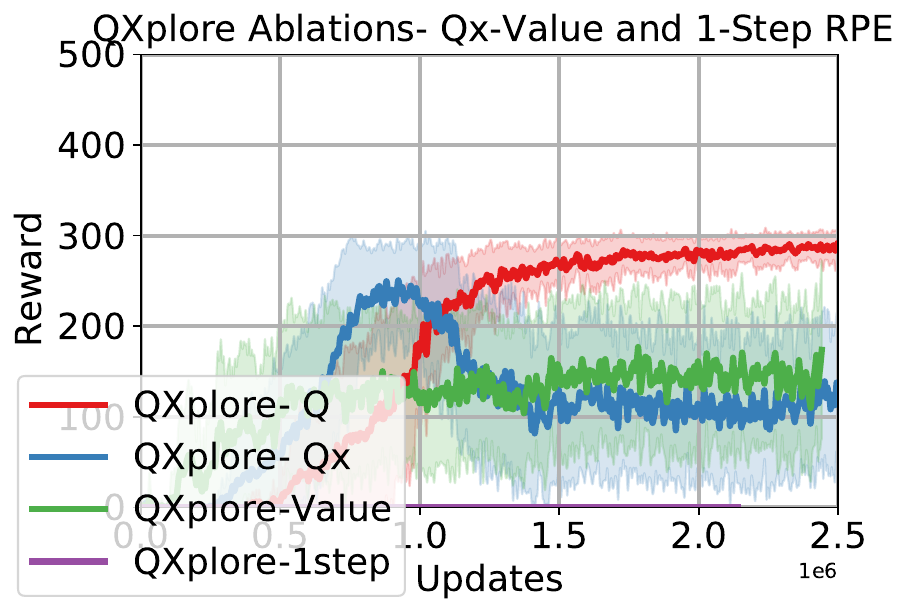}
\includegraphics[width=0.49\columnwidth]{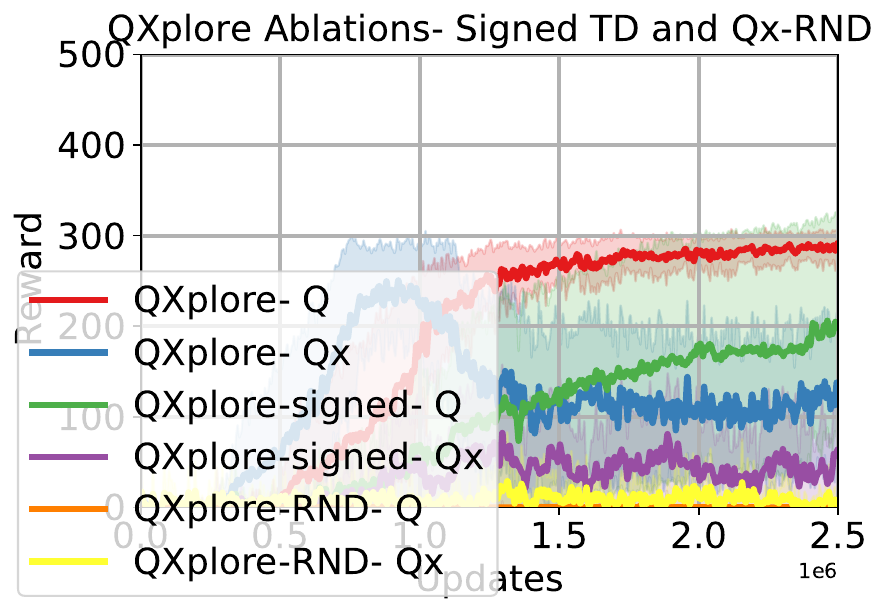}

\caption{Plots showing several ablations of QXplore on \texttt{SparseHalfCheetah}. While several variants are able to learn the task, the full QXplore formulation performs better.}
\label{fig:ablations}
\end{center}
\end{figure}

There are two major features of QXplore that distinguish it from prior work in exploration: the use of a pair of policies that share replay data, and the use of unsigned TD-error to drive exploration. We performed several ablations that assess the contribution of each of these features to our method and confirm their value for exploration. We show the results in Figure \ref{fig:ablations}. We find that the use of separate exploration and exploitation policies along with unsigned TD-error is necessary to obtain good performance, and that ablations of these components either fail to train or substantially reduce performance. We discuss each case in detail below.

\paragraph{Single-Policy QXplore.} First, we tested a single-policy version of QXplore by replacing $Q_{\theta}(s,a)$ with a value function $V_{\theta}(s)$.  We use a value function rather than Q-function in this case to avoid large estimation errors stemming from fully off-policy training. We observe in Figure \ref{fig:ablations} that while the policy is able to find reward quickly and converge faster, the need to satisfy both objectives results in a lower converged reward than the original QXplore method.

\paragraph{1-Step Reward Prediction.} Second, we ran an ablation where we replace $Q_{\theta}(s,a)$ with a function that simply predicts the current $r(s_t,a_t)$. Using reward error instead of a value function in $Q_x$ can still produce the same state novelty fallback behavior in the absence of reward; however, it provides only limited reward-based exploration utility. We evaluate this variant and observe in Figure \ref{fig:ablations} that it fails to sample reward. Reward prediction error is not sufficient to allow strong exploration behavior without some form of lookahead.

\paragraph{QXplore with State Novelty Exploration.} To assess the importance of TD-error specifically in our algorithm, we replaced the TD-error maximization objective of $Q_x$ with the random network prediction error maximization objective of RND, while still performing rollouts of both policies. The results are shown in Figure \ref{fig:ablations}. We observe that while the modified $Q_x$ samples reward, it is too infrequent to guide $Q$ to learn the task. Qualitatively, the modified $Q_x$ function does not display the directional preference in exploration that normal $Q_x$ does once reward is discovered, instead sampling both directions equally.

\paragraph{QXplore with Signed TD-Error Objective.} While we used unsigned TD-error to train $Q_x$, we also tested QXplore using signed TD-error. We used the negative signed TD-error $-\delta_t$ from equation \ref{eq:td_definition} so that better-than-expected rewards result in positive $r_x$ values. The results of this experiment are shown in Figure \ref{fig:ablations}. While this ablation is able to converge and solve the task, the unsigned TD-error performs much better on \texttt{SparseHalfCheetah}, likely due to the extrapolation error described in Figure \ref{fig:0pred} being both positive and negative.

\subsection{Qualitative Behavioral Analysis}
\label{subsec:qualitative}

\begin{figure}[t]
\begin{center}

\includegraphics[width=1.00\columnwidth]{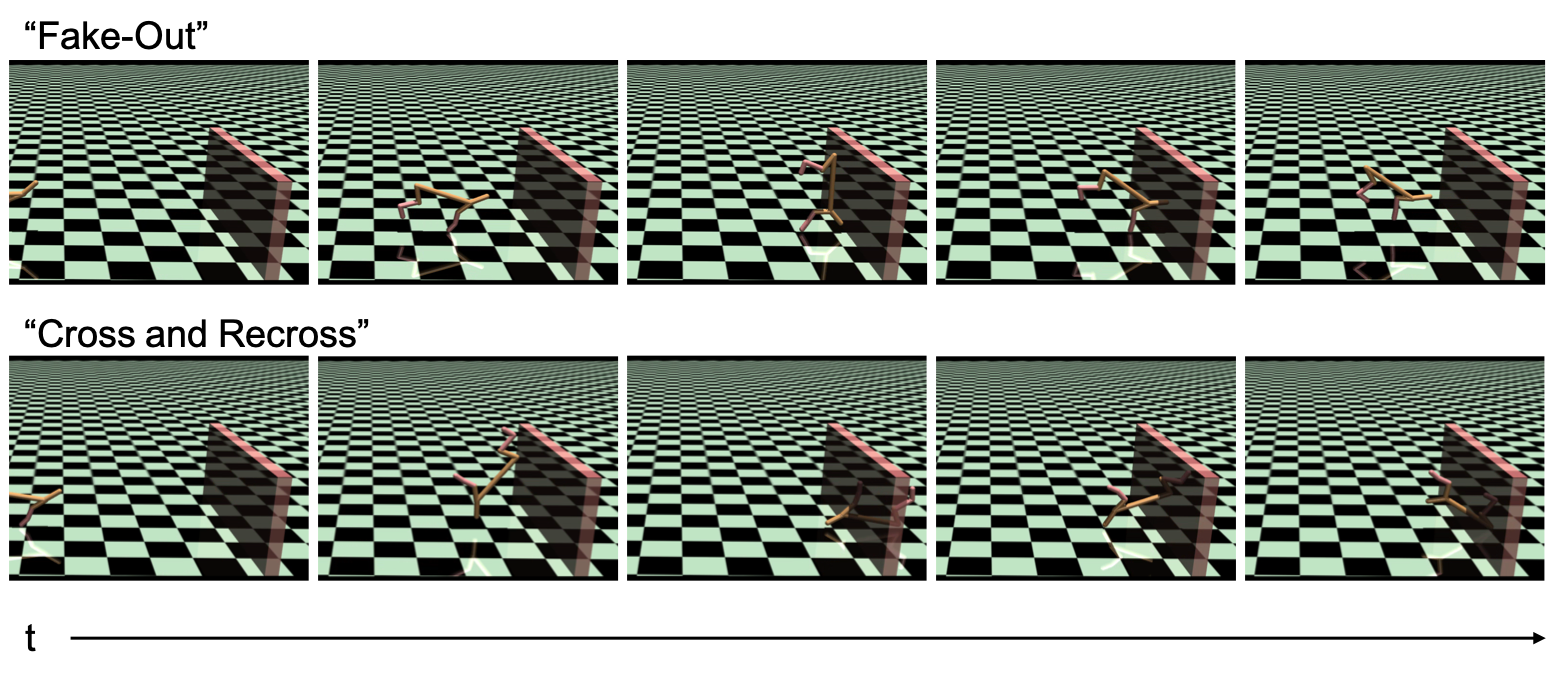}

\caption{Example trajectories showing $Q_x$'s behavior late in training that is distinctive of TD-error maximization. The corresponding $Q$ network reliably achieves reward at this point. In ``fake-out", $Q_x$ approaches the reward threshold and suddenly stops itself. In ``cross and recross", $Q_x$ crosses the reward threshold going forward and then goes backwards through the threshold.}
\label{fig:qualitative}
\end{center}

\end{figure}

Qualitatively, on \texttt{SparseHalfCheetah} we observe interesting behavior from $Q_x$ late in training. After initially converging to obtain high reward, $Q_x$ appears to get ``bored'' and will focus on the reward threshold, stopping short or jumping back and forth across it, which results in reduced reward but higher TD-error. This behavior is distinctive of TD-error seeking over state novelty seeking, as such states are not novel compared to moving past the threshold but do result in higher TD-error. Such behavior from $Q_x$ motivates $Q$ to sample the state space around the reward boundary and thus learn to solve the task. Example sequences of such behaviors are shown in Figure \ref{fig:qualitative}.

\section{Discussion and Conclusions}
\label{sec:conclusion}

Here, we have proposed the use of reward prediction error as an objective for exploration in deep reinforcement learning. We defined a deep RL algorithm, QXplore, using TD-error that is sufficient to discover solutions to multiple types of challenging exploration tasks across multiple domains. We found that QXplore performs well across all exploration task types tested compared to our state novelty baseline, although type-specific algorithms can likely perform better on some types, such as goal-directed exploration.

While QXplore is a general-purpose exploration algorithm that can be applied successfully to many tasks, several limitations remain for TD-error exploration. In the worst-case, TD-error likely performs no better than state novelty for certain ``pure'' exploration tasks, such as exploring a linear chain of states, though with an optimistic prior on the Q-values of unseen states it may perform comparably to state novelty. There also exist adversarial tasks where the unsigned TD-error leads to less efficient exploration compared to other possible policies, such as a task with many states that yield large negative rewards uncorrelated with positive rewards. Combining TD-error exploration with reward exploitation may help in such cases to bias the search. However, balancing the rate at which the TD-error signal disappears for a given state with the reward function's magnitude is critical to get rapid convergence, and more research into such approaches is needed. Lastly, TD-error maximization may result in ``risky'' exploration (in contrast to the ``safe'' TD-minimizing exploration of \citeauthor{Gehring2013smartexploration}) and thus may not be well suited for tasks where failures or negative returns have real-world consequences without additional constraints on the agent's actions, or the use of signed TD-error to avoid trajectories yielding worse-than-expected returns.

We hope that our results can spur more investigation into TD-error-based exploration methods to address some of the outstanding challenges described above, as well as encourage further work on diverse exploration signals in RL and on more general exploration objectives suitable for use on heterogeneous RL tasks.

\bibliographystyle{named}
\bibliography{qxplore}

\end{document}